\documentclass[letterpaper, 10 pt, conference]{ieeeconf} 
\IEEEoverridecommandlockouts

\usepackage{amsmath,mleftright,mathtools}
\usepackage{amssymb}
\usepackage{lipsum}
\usepackage[pdftex, pdfstartview={FitV}, pdfpagelayout={TwoColumnLeft},bookmarksopen=true,plainpages = false, colorlinks=true, linkcolor=black, citecolor = black, urlcolor = black,filecolor=black , pagebackref=false,hypertexnames=false, plainpages=false, pdfpagelabels ]{hyperref}
\usepackage{balance}
\usepackage{xargs}
\usepackage{graphicx}
\usepackage[labelformat=simple]{subcaption}
\DeclareCaptionLabelSeparator{periodspace}{.\quad}
\usepackage[sort,compress]{cite}

\usepackage{tabularx}
\usepackage{bm,etoolbox}
\usepackage[linesnumbered,ruled,vlined]{algorithm2e}
\usepackage[pdftex,dvipsnames]{xcolor}

\DeclareMathOperator*{\argmin}{arg\,min}
\DeclareMathOperator*{\argmax}{arg\,max}
\DeclareGraphicsExtensions{.pdf,.png}
\graphicspath{ {./figs/} }
\DeclarePairedDelimiterX{\infdivx}[2]{(}{)}{%
  #1\;\delimsize\|\;#2%
}

\newcommandx{\unsure}[2][1=]{\todo[linecolor=red,backgroundcolor=red!25,bordercolor=red,#1]{#2}}
\newcommandx{\add}[2][1=]{\todo[linecolor=blue,backgroundcolor=blue!25,bordercolor=blue,#1]{#2}}
\newcommandx{\info}[2][1=]{\todo[linecolor=OliveGreen,backgroundcolor=OliveGreen!25,bordercolor=OliveGreen,#1]{#2}}
\newcommandx{\edit}[2][1=]{\todo[linecolor=Plum,backgroundcolor=Plum!25,bordercolor=Plum,#1]{#2}}
\newcommandx{\hide}[2][1=]{\todo[disable,#1]{#2}}

\SetCommentSty{mycommfont}

\title{\LARGE \bf
NF-iSAM: Incremental Smoothing and Mapping\\ via Normalizing Flows
}

\author{Qiangqiang Huang$^{1,*}$, Can Pu$^{2,*}$, Dehann Fourie$^{1}$, Kasra Khosoussi$^{3}$, Jonathan P. How$^{3}$ and John J. Leonard$^{1}$
\thanks{*Equal contributors}%
\thanks{$^{1}$MIT Computer Science and Artificial Intelligence Laboratory, Cambridge, MA 02139, USA. {\tt\{hqq,dehann,jleonard\}@mit.edu}}%
\thanks{$^{2}$MIT Department of Nuclear Science and Engineering, Cambridge, MA, 02139, USA. {\tt\small {pucan}@mit.edu}}%
\thanks{$^{3}$MIT Department of Aeronautical and Astronautical Engineering, Cambridge, MA, 02139, USA. {\tt\small{kasra,jhow}@mit.edu}}%
\thanks{Research supported by ONR grant N00014-18-1-2832 and ONR MURI grant N00014-19-1-2571.}
}

\begin{document}
\maketitle
\thispagestyle{empty}
\pagestyle{empty}
\begin{abstract}
This paper presents a novel non-Gaussian inference algorithm, Normalizing Flow iSAM (NF-iSAM), for solving SLAM problems with non-Gaussian factors and/or non-linear measurement models. NF-iSAM exploits the expressive power of neural networks, and trains normalizing flows to draw samples from the joint posterior of non-Gaussian factor graphs. By leveraging the Bayes tree, NF-iSAM is able to exploit the sparsity structure of SLAM, thus enabling efficient incremental updates similar to iSAM2, albeit in the more challenging non-Gaussian setting. We demonstrate the performance of NF-iSAM and compare it against the state-of-the-art algorithms such as iSAM2 (Gaussian) and mm-iSAM (non-Gaussian) in synthetic and real range-only SLAM datasets.
\end{abstract}
\section{Introduction}

Simultaneous localization and mapping (SLAM) is usually formulated as a Bayesian inference problem \cite{dellaert2017factor}.  The state-of-the-art SLAM algorithms such as iSAM2~\cite{kaess2012isam2} seek the maximum \emph{a posteriori} (MAP) estimate.
Under the assumption of Gaussian measurement noise and priors, these methods can also provide a Gaussian approximation to the posterior. However, the posterior distribution in real-world SLAM problems is almost always non-Gaussian and may have multiple modes. This is in part due to non-linear measurement models and non-Gaussian factors~\cite{fourie2016nonparametric}. Examples include real-world scenarios involving range measurements \cite{fourie2017multi}, pose transformations on the special Euclidean group \cite{long2013banana}, multi-modal data association \cite{doherty2020probabilistic}, and odometry slip\slash grip error modelling \cite{olson2013inference}. Therefore, Gaussian (or any unimodal) approximation is \emph{inherently} incapable of capturing critical information about the uncertainty in SLAM, which is essential for safe navigation. The main challenge is that \emph{exact} inference is in general computationally intractable. This highlights the need for efficient  \emph{approximate} inference methods that can adequately capture and represent uncertainty in SLAM.

The computational cost of traditional general-purpose methods, like Markov chain Monte Carlo (MCMC) or nested sampling, is usually too high for high-dimensional factor graphs that arise in SLAM \cite{kaess2005markov,torma2010markov,skilling2006nested}. The efficiency of the state-of-the-art SLAM algorithms such as iSAM2 lies in exploiting the sparsity structure of SLAM via the Bayes tree. The Bayes tree converts a high-dimensional cyclic factor graph into an acyclic directed graph using an elimination game~\cite{Heggernes1996finding} and max-cardinality search~\cite{Tarjan1984simple}. In essence, the Bayes tree decomposes the high-dimensional inference problem into low-dimensional inference problems on cliques.
Recent extensions of iSAM2, such as mm-iSAM~\cite{fourie2017multi} and MH-iSAM2~\cite{hsiao2019mh}, all take advantage of the acyclic Bayes tree
to exploit the sparsity pattern of the original high-dimensional factor graphs~\cite{kaess2010bayes, Fourie2020wafr}.  These algorithms, however, can only infer the \emph{marginal} posterior distribution (mm-iSAM) or are limited to certain sources of non-Gaussianity (MH-iSAM2). 

\begin{figure}[t] 
 	\centering
 	\includegraphics[width=0.75\columnwidth]{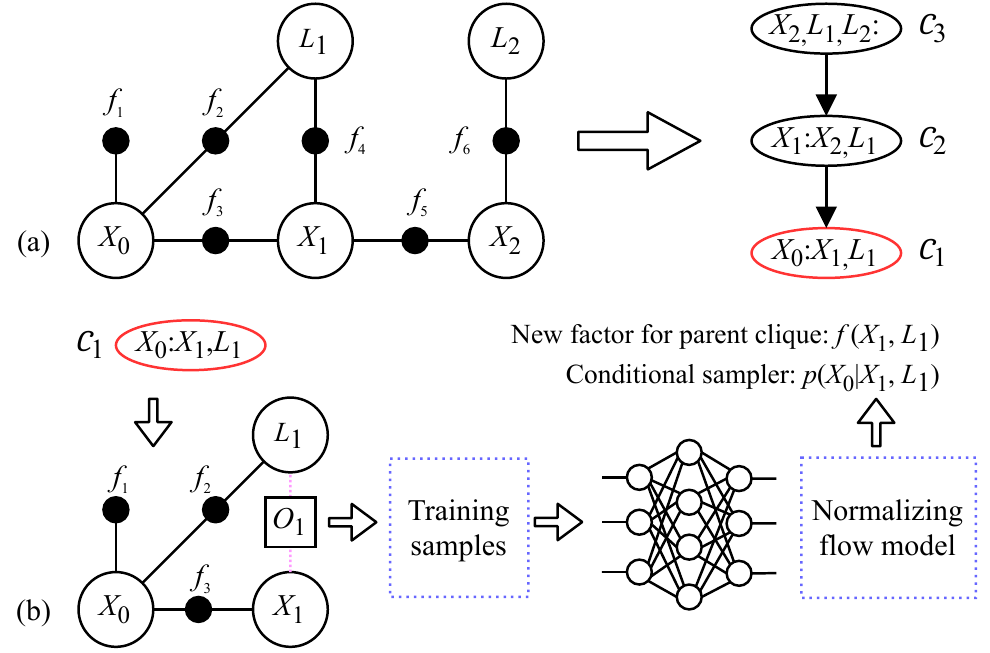}
     \caption{Illustration of core steps in NF-iSAM: (a) conversion from a factor graph to the Bayes tree with elimination ordering $(X_0, X_1, X_2, L_1, L_2)$ and (b) construction of clique conditional sampler via normalizing flows. The colon in a Bayes tree node splits frontal and separator variables. The normalizing flow model is learnt from training samples by neural networks. In factors of Bayes tree node $\mathcal{C}_1$, the measured value of $f_4$ is relaxed to forecast observation variable $O_1$ for rapidly drawing training samples.}
     \vspace{-0.5 cm}
     \label{fig: fig 1 illustration}
\end{figure}

In this paper, we present a novel general solution, Normalizing Flow iSAM (NF-iSAM), to infer the joint posterior distribution of general non-Gaussian SLAM problems using the Bayes tree and the normalizing flow model \cite{rezende2015variational}. We present experimental results from a synthetic dataset and a real dataset for range-only SLAM.
Key features of NF-iSAM include the following:
\begin{enumerate}
    \item   NF-iSAM extends normalizing flows from low-dimensional inference to high-dimensional cyclic factor graphs by exploiting the sparsity structure encoded in the Bayes tree to efficiently perform inference incrementally.
    \item   NF-iSAM extends iSAM2 to non-Gaussian factor graphs by exploiting the expressive power of normalizing flows and neural networks on probabilistic modeling.
    \item   Unlike the state-of-the-art non-Gaussian factor graph inference algorithm, mm-iSAM~\cite{fourie2016nonparametric,Fourie2019iros}, which can only generate samples from the \emph{marginal} posterior distributions, NF-iSAM is able to draw samples from the \emph{joint} posterior distribution.
\end{enumerate}

\section{Related Work}

Existing methods for factor graph inference can be categorized into two classes, namely parametric and non-parametric solutions. 
The state-of-the-art optimization-based solutions to SLAM, such as iSAM2~\cite{kaess2012isam2}, are MAP point estimators that approximate the posterior distribution by a single, parametric Gaussian model~\cite{dellaert2017factor}, thus, they are not general solutions to non-Gaussian posteriors.
iSAM2 has also been extended to multi-hypothesis problems consisting of Gaussian mixture factors. The main challenge in such problems is the exponential number of Gaussian modes. The max-mixture \cite{olson2013inference} approach avoids the exponential complexity by selecting a highly probable component from all modes. However, uncertainty information of all other modes cannot be recovered in the max-mixture solution. MH-iSAM2~\cite{hsiao2019mh} aims to retain dominant hypotheses by pruning less likely hypotheses. While the multi-hypothesis approach can address uncertain data association and ambiguous loop closures, it cannot explicitly tackle more general non-Gaussianity such as range-only measurements.

Nonparametric models are able to capture some non-Gaussian posterior densities. These methods use sampling techniques, such as particle filters or MCMC \cite{montemerlo2003fastslam,kaess2005markov,torma2010markov}. The most well-known nonparametric SLAM algorithm is FastSLAM 2.0 \cite{montemerlo2003fastslam}. Sequential Monte Carlo methods such as FastSLAM suffer from the so-called particle depletion and degeneracy problems. Moreover, for each sample of robot's trajectory, FastSLAM uses extended Kalman filters to obtain a Gaussian approximation to the posterior distribution of the map. A more recent method, multimodal-iSAM (mm-iSAM)~\cite{fourie2016nonparametric,Fourie2019iros}, leverages the Bayes tree \cite{kaess2010bayes} to solve SLAM problems with a variety of non-Gaussian error sources~\cite{fourie2017multi, Fourie2020iros}.
mm-iSAM uses nested Gibbs sampling, derived from nonparametric belief propagation~\cite{sudderth2003nonparametric}, to approximate the \emph{marginal} belief of each clique. mm-iSAM is unable to generate samples from the \emph{joint} posterior distribution.

In the machine learning community, non-Gaussian inference has also drawn researchers' interest, however most works focus on low-dimensional problems instead of general factor graphs. 
Kernel embedding is a tool to represent non-Gaussian densities \cite{smola2007hilbert, song2013kernel}, and can be applied to hidden Markov models \cite{fukumizu2013kernel, kanagawa2014monte, gebhardt2017kernel} and trees \cite{song2011kernel}. 
Although this method has been successfully applied to a number of robotic problems \cite{mccalman2013multi, kim2018imitation}, converting an embedding back to a density is computationally challenging\cite{scholkopf2002learning}.
The nonparanormal model \cite{lafferty2012sparse}, also known as the Gaussian copula, is an extension to the Gaussian distribution. 
It represents a density that can be transformed to a Gaussian after marginally applying an increasing map along each dimension, and has been applied to trees \cite{elidan2012nonparanormal}. Recent work uses a copula as the proposal distribution for particle filters~\cite{martin2020variational}, but is not yet tested at the scale of SLAM problems.

Another class of algorithms draws samples from a non-Gaussian target distribution by transforming samples from a simple reference distribution. 
These methods are known as transport maps \cite{el2012bayesian}, or normalizing flows \cite{rezende2015variational} when the reference distribution is Gaussian.
Although they have shown good performance in noise modeling \cite{abdelhamed2019noise} and reinforcement learning \cite{mazoure2020leveraging}, research on high-dimensional graphical models is limited. 
Specifically, existing solutions are still limited to problems with special structures, such as a hidden Markov model or data assimilation  \cite{spantini2018inference, spantini2019coupling}. In this work, we leverage the Bayes tree to extend normalizing flow from a low-dimensional solution to high-dimensional cyclic factor graphs that arise in SLAM. The proposed algorithm, NF-iSAM, is able to draw samples from the non-Gaussian joint posterior distribution in SLAM problems.

\section{Factor Graphs and the Bayes Tree}
\label{sec:factor graph and Bayes tree}
A factor graph consists of variables and factors. 
Given a variable elimination ordering, a factor graph can be converted to a Bayes tree by Algorithm~2 and~3 in \cite{kaess2012isam2}. 
Nodes on the Bayes tree represent cliques of variables as shown in Fig.\ \ref{fig: fig 1 illustration}(a). 
Variables on a clique shared with its parent clique are called separator variables while the remainder are frontal variables. 
The Bayes tree represents the following factorization of the posterior
\begin{equation}
    p(\Theta|z)=\prod_{\mathcal{C}\in\mathbf{C}} p(F_\mathcal{C}|S_\mathcal{C},z)=\prod_{\mathcal{C}\in\mathbf{C}} p(F_\mathcal{C}|S_\mathcal{C},z_\mathcal{C}),\label{eqn: Bayes tree factorization}
\end{equation}
where $\Theta$ is a high-dimensional random variable consisting of all poses and landmark locations, $z$ denotes all of the robot's observations, $\mathbf{C}$ is the collection of cliques, $F_\mathcal{C}$ denotes the set of frontal variables at clique $\mathcal{C}$, $S_\mathcal{C}$ denotes separator variables, and $z_\mathcal{C}$ denotes the set of observations in and below clique $\mathcal{C}$ on the Bayes tree.

If, for every clique $\mathcal{C}$ and any $S_\mathcal{C}=s_\mathcal{C}$, we could construct a conditional sampler that can efficiently draw independent samples of $F_\mathcal{C}$ from $p(F_\mathcal{C}|S_\mathcal{C}=s_\mathcal{C},z_\mathcal{C})$, then we would be able to sample the joint posterior from the root to the leaves of the Bayes tree. In the Bayes tree construction process of clique $\mathcal{C}$, the clique joint density $p(S_\mathcal{C},F_\mathcal{C}|z_\mathcal{C})$ is refactored as product of the conditional density $p(F_\mathcal{C}|S_\mathcal{C},z_\mathcal{C})$ and the separator marginal density $p(S_\mathcal{C}|z_\mathcal{C})$. $p(S_\mathcal{C}|z_\mathcal{C})$ will be attached to its parent clique as new factors in the elimination process, thus, it will be used in following computation and we also need to characterize it.

\section{SLAM via normalizing flows}
\label{sec: method}
In this section, we first briefly review normalizing flows (Section \ref{sec: normalizing flows}). We then present our novel technique for building samplers for $p(F_\mathcal{C}|S_\mathcal{C},z_\mathcal{C})$ and $p(S_\mathcal{C}|z_\mathcal{C})$ via normalizing flows in Section \ref{sec: clique conditional sampler}. Lastly, in Section \ref{sec: incremental update}, we describe our \emph{incremental} inference approach that generates joint posterior samples.
\subsection{Normalizing Flows}
\label{sec: normalizing flows}
A normalizing flow is the transformation $T$ from a target random variable $X\in \mathbb{R}^{D}$ to another variable $Y$ that follows a reference distribution $q_{Y}(y)$ which is standard Gaussian. 
The transformation here is a lower-triangular map:
\begin{align}
T(x)=
\begin{bmatrix*}[l]
  T_1(x_1)\\
  T_2(x_1,x_2)\\
  \vdots\\
  T_D(x_1,x_2,\hdots,x_D)\\
\end{bmatrix*}
=
\begin{bmatrix*}[c]
  y_1\\
  y_2\\
  \vdots\\
  y_D\\
\end{bmatrix*}
=
y,\label{eq: change of variables}
\end{align}
where each row $T_d$ is differentiable, bijective, and increasing with respect to $x_d$ \cite{rezende2015variational,durkan2019neural,jaini2019sum,papamakarios2019normalizing,rezende2020normalizing}. Therefore, if the transformation $T$ is known, we can easily draw samples from $x \sim p_{X}$ by
\begin{enumerate}
    \item   drawing samples $y\sim q_{Y}$ and
    \item   solving the inverse transformation problem
    \begin{align}
    x
    =
    T^{-1}(y)
    =
    \begin{bmatrix*}[l]
      T_{1}^{-1}(y_1)\\
      T_{2}^{-1}(y_2;x_1)\\
      \vdots\\
      T_{D}^{-1}(y_D;x_1,x_2,\hdots,x_{D-1})
    \end{bmatrix*}.\label{eqn:inverse map}
    \end{align}
\end{enumerate}
Since in \eqref{eqn:inverse map} one first solves for $\{x_i\}_{i=1}^{d-1}$ before solving for $x_d$, finding $x_d$ requires inverting a one-dimensional function given $y_d$ and $\{x_i\}_{i=1}^{d-1}$. In general, the transformation between random variables of two distributions is not unique. It has been proven, however, that triangular maps to a standard Gaussian exist and are unique for any non-vanishing densities \cite{carlier2010knothe,bonnotte2013knothe,bogachev2005triangular,villani2008optimal}. The triangular map has also been exploited in prior work on Bayesian inference and sampling \cite{el2012bayesian, spantini2018inference, parno2018transport}. An important property of the triangular map is that $T_{d}$ encodes the conditional probability $p(x_d|x_{d-1},...,x_{1})$ \cite{jaini2019sum}. For examples, function $T_1(\cdot)$ transforms marginal sample $x_1\sim p(x_1)$ to a sample $y_1$ that follows a one-dimensional standard Gaussian distribution $q_1$. For $d\neq 1$, if a sample $x_d$ follows the conditional distribution $p(x_d|X_1=x_1, X_2=x_2, \ldots, X_{d-1}=x_{d-1})$, then conditional normalizing flow  $T_d(x_1,x_2,\ldots, x_{d-1}, \cdot)$ transforms $x_d$ to a one-dimensional standard Gaussian sample $y_d\sim q_d$. This property is important and will be used to build the desired clique conditional sampler in Section \ref{sec: clique conditional sampler}.

Our goal here is to parameterize the triangular map and solve for it. $T_d$ can be expressed as a one-dimensional function of $x_d, T_d(x_d;c_d(x_1,x_2,...,x_{d-1}))$,
where $c_d(x_1,x_2,...,x_{d-1})$ is named a conditioner whose output is a set of parameters fixing the one-dimensional function. For example, sum-of-square polynomials and splines can be used to parameterize the function $T_d$, so  the outputs of their conditioners will be polynomial coefficients or spline segments, respectively \cite{jaini2019sum, durkan2019neural}.

Given a family of functions $\mathcal{F}$ for parameterization, and $n$ training samples of $x$, we seek the triangular map minimizing the Kullback–Leibler (KL) divergence between $p(x)$ and $p_T(x)$, where $p_T(x)$ denotes the density transformed from $q(y)$ by $T^{-1}$ for a triangular map $T$. For any $T$, by change of variables, we have $p_T(x)=q(T(x)) \,\, |{T'}(x)|$,
where $|{T'}(x)|$ is the determinant of the Jacobian. Therefore, with the $n$ training samples, an optimal triangular map $T^\star \in \argmin_{T\in\mathcal{F}}\text{KL}\left({p(x)}\|{p_T(x)}\right)$ is given by
\begin{align}
    T^{\star} &\in \argmin_{T\in\mathcal{F}} \int_{x}p(x) \log{\frac{p(x)}{p_T(x)}}\,\text{d}x,\\
    &= \argmax_{T\in\mathcal{F}} \int_{x}p(x) \left[ \log{q(T(x))}+\log{|{T'}(x)|}\right]\,\text{d}x,\\
    &\approx \argmax_{T\in\mathcal{F}} \sum_{k=1}^{n} \left[ \log{q(T(x_{(k)}))}+\log{|{T'}(x_{(k)})|} \right],\label{eqn:MC KL}
\end{align}
where the last step uses samples from $p$ and Monte Carlo integration as approximation. 

\begin{figure}[t]\vspace*{3mm}
	\centering
	\includegraphics[width=.95\columnwidth]{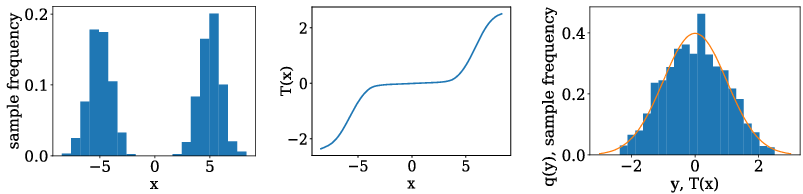}
	\vspace*{-0.05cm}
    \caption{A one-dimensional example of normalizing flow: histogram of sample $x$ (left), transformation function $T(x)$ (middle), and histogram of transformed samples and reference variable $y\sim N(0,1)$ (right).}
    \label{fig:1d normalizing flow}
    \vspace{-0.6cm}
\end{figure}

Considering the flexibility of splines, we use rational-quadratic splines to parameterize $T_d$ \cite{durkan2019neural}. The parameters are coordinates of spline knots on the $x$-$y$ plane and spline derivatives at those knots. For a univariate transformation, e.g., $T_1$, its conditioner does not depend on $x$ and just outputs constant knot parameters. We can simply optimize over those knot parameters to find a univariate transformation minimizing the KL divergence. Fig.~\ref{fig:1d normalizing flow} shows such a one-dimensional transformation that maps raw samples to standard Gaussian samples. However, for higher dimensional functions, e.g. $T_{d}$ with $d>1$, those knot parameters have to depend on $\{x_i\}_{i=1}^{d-1}$. Exploiting the expressive power of neural networks, we use fully connected neural networks to model those conditioners; see  \cite[Sec.\ 3]{durkan2019neural} for details of the parameterization. Our implementation is adapted from \cite{normalizing-flows}.
A usual routine before training is standardizing raw samples by their means and standard deviations to regularize unbounded large values \cite{ioffe2015batch}. Samples of orientation variables are transformed to $[-\pi,\pi]$ before being standardized. This standardizing step is equivalent to an affine transformation which makes training more efficient and does not alter the problem nor affect the non-Gaussianity in the raw samples. When the training is finished, the resulting triangular map will be transformed back to the space of raw samples by the inverse of the affine transformation.

\subsection{Clique Conditional Samplers via Normalizing Flows}
\label{sec: clique conditional sampler}
As suggested in Section \ref{sec: normalizing flows}, we can learn the normalizing flow
\begin{equation}
    T_{\mathcal{C}}(S_\mathcal{C}, F_\mathcal{C})=
    \begin{bmatrix*}[l]
        T_{S_\mathcal{C}}(S_\mathcal{C})\\
        T_{F_\mathcal{C}}(S_\mathcal{C},F_\mathcal{C})\\
    \end{bmatrix*}
    \label{eq: conditional normalizing flows without observations}
\end{equation}
for density $p(S_\mathcal{C},F_\mathcal{C}|z_\mathcal{C})$ on clique $\mathcal{C}$ if we have training samples from $p(S_\mathcal{C}, F_\mathcal{C}|z_\mathcal{C})$. $T_{S_\mathcal{C}}$ is the normalizing flow for separator marginal $p(S_\mathcal{C}|z_\mathcal{C})$, and $T_{F_\mathcal{C}}$ is the conditional normalizing flows for $p(F_\mathcal{C}|S_\mathcal{C},z_\mathcal{C})$. There are many well-developed off-the-shelf implementations of MCMC sampling such as {PyMC3} \cite{salvatier2016probabilistic} or nested sampling such as {dynesty} \cite{speagle2020dynesty}. However, even though the dimension of a clique is much smaller than that of the entire factor graph, those packages are still too slow for obtaining training samples on a clique in real-time for robotics applications since they involve iterative evaluation of densities.

Inspired by the so-called forecast-analysis scenario in hidden Markov models \cite{spantini2019coupling}, we propose the following alternative strategy:
\begin{enumerate}
    \item   Draw training samples from another density $\widetilde{p}$, where sampling is efficient.
    \item   Train normalizing flow $\widetilde{T}$ for $\widetilde{p}$, and retrieve $T$ for $p(S_\mathcal{C},F_\mathcal{C}|z_\mathcal{C})$.
\end{enumerate}

In step 1, we sample from density $p(O_\mathcal{C},S_\mathcal{C},F_\mathcal{C}|z_\mathcal{C}')$, where $z_\mathcal{C}'=z_\mathcal{C}\backslash O_\mathcal{C}$. We can select a set of likelihood factors, whose measurement variables are $O_\mathcal{C}$, that breaks the clique into a forest, where each tree has a node with prior samples. For example, in Fig. \ref{fig: fig 1 illustration}(b), there is only one tree with variables $X_0$, $X_1$ and $L_1$, $f_4$ is the factor that breaks the clique into a tree, and $O_1$ is its measurement variable. With the knowledge of measurement models, we can efficiently sample these trees using ancestral sampling \cite{bishop2006pattern}. Starting from these samples which follow $p(S_\mathcal{C}, F_\mathcal{C}|z_\mathcal{C}')$, we generate samples for measurements $O_\mathcal{C}$ using measurement models. For example, in Fig.~\ref{fig: fig 1 illustration}(b), we simulate forecast observation $O_1$ between $X_1$ and $L_1$ samples by $f_4$ noise. Thereby, a sample is drawn without the need of iterative procedures (see Algorithm \ref{algo: clique sampler}).
This strategy works for any clique $\mathcal{C}$ when $\mathcal{C}$ has no node connected to two prior factors simultaneously, and this is ensured by the elimination ordering.

In step 2, we use training samples from $p(O_\mathcal{C},S_\mathcal{C},F_\mathcal{C}|z_{\mathcal{C}}^{\prime})$ to obtain the normalizing flow $\widetilde{T}_\mathcal{C}$ for $p(S_\mathcal{C}|z_\mathcal{C})$ and $p(F_\mathcal{C}|S_\mathcal{C},z_\mathcal{C})$. According to Section \ref{sec: normalizing flows}, by ordering variables in $\widetilde{T}_\mathcal{C}$ according to $(O_\mathcal{C}, S_\mathcal{C}, F_\mathcal{C})$, we get the triangular map
\begin{equation}
    \widetilde{T}_\mathcal{C}(O_\mathcal{C}, S_\mathcal{C}, F_\mathcal{C})=
    \begin{bmatrix*}[l]
        \widetilde{T}_{O_\mathcal{C}}(O_\mathcal{C})\\
        \widetilde{T}_{S_\mathcal{C}}(O_\mathcal{C},S_\mathcal{C})\\
        \widetilde{T}_{F_\mathcal{C}}(O_\mathcal{C},S_\mathcal{C},F_\mathcal{C})
    \end{bmatrix*}.
    \label{eq: conditional normalizing flows with observations}
\end{equation}
When we fix $O_\mathcal{C}$ to its measured value $o_\mathcal{C}$, $\widetilde{T}_{S_\mathcal{C}}(o_\mathcal{C},\cdot)$ gives the normalizing flow for separator $S_\mathcal{C}\sim p(S_\mathcal{C}|z_\mathcal{C})$, and $\widetilde{T}_{F_\mathcal{C}}(o_\mathcal{C},s_\mathcal{C},\cdot)$ gives the conditional normalizing flow for $F_\mathcal{C}\sim p(F_\mathcal{C}|S_\mathcal{C}=s_\mathcal{C},z_\mathcal{C})$ (see Algorithm \ref{algo: clique normalizing flows}).
Therefore, we can retrieve $T_\mathcal{C}$ from $\widetilde{T}_\mathcal{C}$:
\begin{equation}
    T_\mathcal{C}\left(S_\mathcal{C},F_\mathcal{C}\right)
    =
    \begin{bmatrix*}[l]
        \widetilde{T}_{S_\mathcal{C}}\left(O_\mathcal{C}=o_\mathcal{C},S_\mathcal{C}\right) \\
        \widetilde{T}_{F_\mathcal{C}}\left(O_\mathcal{C}=o_\mathcal{C},S_\mathcal{C},F_\mathcal{C}\right)
    \end{bmatrix*}.
    \label{eqn: T tilde to T}
\end{equation}

\subsection{Incremental Inference on Bayes Tree}
\label{sec: incremental update}
Our inference will start from the leaf clique ${\mathcal{C}_\text{L}}$. 
By Section \ref{sec: clique conditional sampler}, we can learn samplers for $p(S_{\mathcal{C}_\text{L}}|z_{\mathcal{C}_\text{L}})$ and $p(F_{\mathcal{C}_\text{L}}|S_{\mathcal{C}_\text{L}},z_{\mathcal{C}_\text{L}})$. Then $p(S_{\mathcal{C}_\text{L}}|z_{\mathcal{C}_\text{L}})$ will be passed to the parent clique as its new prior factor as shown in Fig.\ \ref{fig: fig 1 illustration}(b); $p(F_{\mathcal{C}_\text{L}}|S_{\mathcal{C}_\text{L}},z_{\mathcal{C}_\text{L}})$ will be saved on clique ${\mathcal{C}_\text{L}}$ for sampling the joint posterior density later. Our algorithm learns all clique conditional samplers during a single upward pass on the Bayes tree.

\setlength{\floatsep}{0.0 cm}
\begin{algorithm}[t]
\fontsize{9pt}{9pt}\selectfont
\DontPrintSemicolon 
\KwIn{Prior $\mathcal{P}$ and likelihood $\mathcal{L}$ factors in the clique}
\KwOut{Training samples}
Sample variables in $\mathcal{P}$\;

\While{$\mathcal{L} \neq\emptyset$}{
    $f \gets \mathcal{L}\mathsf{.pop()}$\;
    \uIf{\text{\normalfont all variables in $f$ sampled}}{
      Sample forecast observations in $f$\;
    }
    \uElseIf{\text{\normalfont only one variable in $f$ unsampled}}{
      Sample that variable\;
    }
    \Else{
      $\mathcal{L}\mathsf{.push(}f\mathsf{)}$\;
    }
}
\Return{\text{\normalfont All training samples}}\;
\caption{{CliqueTrainingSampler}}
\label{algo: clique sampler}
\end{algorithm}

\setlength\floatsep{.0 cm}
\begin{algorithm}
\fontsize{9pt}{9pt}\selectfont
\DontPrintSemicolon 
\KwIn{Training samples and realized observations $o$}
\KwOut{Conditional sampler and separator factor}
Rearrange training samples to the order of observation ($O$), separator ($S$), and frontal variables ($F$)\;
Find $\widetilde{T}$ in \eqref{eq: conditional normalizing flows with observations} by minimizing the KL divergence according to \eqref{eqn:MC KL} using the training samples\;
$T(S,F) \gets \widetilde{T}(O=o,S,F)$ \tcp*{fix observations in (\ref{eqn: T tilde to T})}
$T_{S},T_{F} \gets $ partition $T(S,F)$ following (\ref{eq: conditional normalizing flows without observations})\;
Obtain samplers of $p(F|S)$, $p(S)$ from  $T_{S}$ and $T_{F}$ by (\ref{eqn:inverse map})\;
\Return{\text{\normalfont Samplers of} $p(F|S), p(S)$}\;
\caption{{ConditionalSamplerTrainer}}
\label{algo: clique normalizing flows}
\end{algorithm}
\setlength\textfloatsep{.0 cm}
\begin{algorithm}[!ht]
\fontsize{9pt}{9pt}\selectfont
\DontPrintSemicolon 
\KwIn{New factors $\mathcal{F}$, factor graph $\mathcal{G}$, ordering $\mathcal{O}$}
\KwOut{Samples of the joint posterior distribution}
$\mathcal{T} \gets \mathcal{G} \mathsf{.update(}\mathcal{F},\mathcal{O}\mathsf{)}$ \tcp*{update the Bayes tree}
$\mathcal{T}_s \gets \mathcal{T}\mathsf{.extract(}\mathcal{F},\mathcal{O}\mathsf{)}$ \tcp*{extract the affected sub-tree of $\mathcal{T}$}
  \For{\text{\normalfont clique} $\mathcal{C}$ \text{\normalfont in leaf-to-root traverse of} $\mathcal{T}_s$}{
    Training samples $\mathbf{x} \gets$ CliqueTrainingSampler($\mathcal{C}$)\;
    $p(F_\mathcal{C}|S_\mathcal{C}), p(S_\mathcal{C})\gets$
    ConditionalSamplerTrainer($\mathbf{x}$, $o_{\mathcal{C}}$)\;
    Append $p(S_\mathcal{C})$ to the parent clique as a factor\;
    }
$\mathcal{D}\gets \{\}$ \tcp*{initialize an empty dictionary for posterior samples}
  \For{\text{\normalfont clique} $\mathcal{C}$ \text{\normalfont in root-to-leaf traverse of} $\mathcal{T}$}{
  $p(F_\mathcal{C}|S_\mathcal{C}) \gets$ retrieve the conditional sampler in $\mathcal{C}$\;
  $s \gets \mathcal{D}[S_\mathcal{C}]$ \tcp*{retrieve samples of separator variables}
  $\mathcal{D}[F_\mathcal{C}] \gets$ draw samples from $p(F_{\mathcal{C}}|S_{\mathcal{C}}=s)$ using (\ref{eqn:inverse map})\;
}
\Return{$\mathcal{D}$}\;
\caption{{NF-iSAM}}
\label{algo: incremental infernce}
\end{algorithm}

Once all cliques have learned their conditional samplers, we start sampling the joint posterior $p(\Theta|z)$. 
At the root clique $\mathcal{C}_\text{R}$, we use the sampler for $p(F_{\mathcal{C}_\text{R}}|z_{\mathcal{C}_\text{R}})$ to generate samples for $F_{\mathcal{C}_\text{R}}$, which includes separator samples for its child cliques. 
For the downward pass from root to leaves, clique ${\mathcal{C}_k}$ has its separator samples $s_{\mathcal{C}_k}$ shared from its parent clique. We fix $S_{\mathcal{C}_k}$ to $s_{\mathcal{C}_k}$, and use the conditional sampler for $p(F_{\mathcal{C}_k}|S_{\mathcal{C}_k}=s_{\mathcal{C}_k},z_{\mathcal{C}_k})$ to draw samples for $F_{\mathcal{C}_k}$. 
Once all cliques are visited, the algorithm terminates and we have samples from the joint posterior density. 
Compared to learning the conditional samplers, the computational cost of the downward pass is minimal.

When performing incremental updates, we do not need to recompute normalizing flows for all cliques of the Bayes tree. 
Every time a new factor is added into the factor graph, only a part of the old Bayes tree needs to be updated \cite{kaess2012isam2, Fourie2020wafr}. 
We take out the affected part of the factor graph and form a sub-tree from the Bayes tree construction process. 
The upward pass starts from leaves of the sub-tree instead of the entire Bayes tree. 
Normalizing flows for cliques outside the sub-tree are not changed and can be reused directly. 
Thus, the computational cost for training clique conditional samplers depends only on the sub-tree instead of the entire problem. To draw samples from the full joint posterior density, the downward pass still needs to visit all cliques. However, as mentioned above, the computational cost for the downward sampling pass is much lower than that for training normalizing flows. The detailed algorithm is summarized in Algorithm \ref{algo: incremental infernce}.

\section{Results}
\subsection{A Synthetic Dataset}
In our first experiment, we use a small-scale synthetic range-only SLAM dataset where we can explicitly show  posterior samples. Since the dimension of this problem is not too high, it is possible to obtain a reference solution (i.e., samples from the "true" posterior) by a nested sampling package, {dynesty} \cite{speagle2020dynesty}.
The objective in this experiment is to demonstrate that the algorithm correctly captures the non-Gaussian posterior. 
Comparisons are made with the sate-of-the-art methods iSAM2 and mm-iSAM, which are implemented in {GTSAM} and {Caesar.jl}, respectively \cite{dellaert2012factor,caesarjl}.

We assume there is a mobile robot operating on the $x$-$y$ plane. At each time step, the robot moves forward, rotates in-place, and then takes a range measurement with known data association. Gaussian noises are used to perturb the ground truth data for synthesizing pose transformation and range measurements. Note that since both the odometry measurements (pose transformations) and range measurements are nonlinear, the resulting posterior is non-Gaussian with multiple modes.

\begin{figure}[t]\vspace*{3mm}
\centering
\begin{subfigure}{.45\linewidth}
  \centering
  \includegraphics[width=1.0\linewidth]{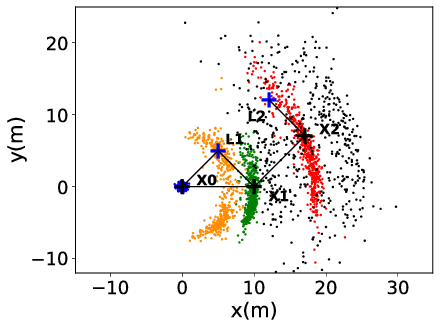}
  \vspace*{-15pt}
  \subcaption{Step 2 (NF-iSAM)}
  \label{fig: step 2 flow}
\end{subfigure}%
\begin{subfigure}{.45\linewidth}
  \centering
  \includegraphics[width=1.0\linewidth]{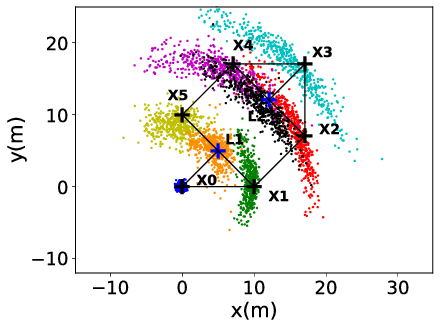}
  \vspace*{-12pt}
  \subcaption{Step 5 (NF-iSAM)}
  \label{fig: step 5 flow}
\end{subfigure}
\newline
\begin{subfigure}{.45\linewidth}
  \centering
  \vspace*{8pt}
  \includegraphics[width=1.0\linewidth]{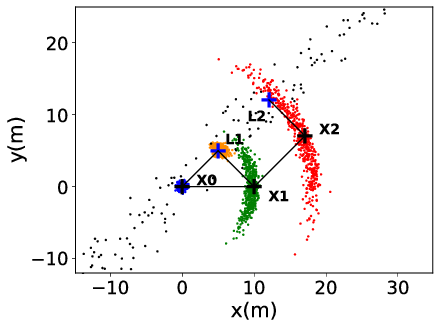}
  \vspace*{-12pt}
  \subcaption{Step 2 ({iSAM2})}
  \label{fig: step 2 GTSAM}
\end{subfigure}
\begin{subfigure}{.45\linewidth}
  \centering
    \vspace*{8pt}
  \includegraphics[width=1.0\linewidth]{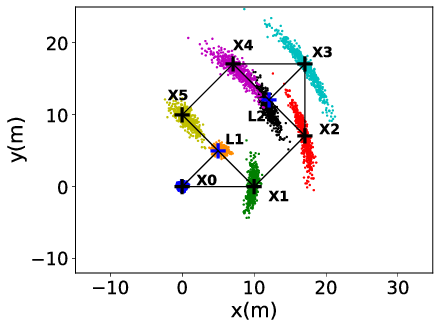}
  \vspace*{-12pt}
  \subcaption{Step 5 ({iSAM2})}
  \label{fig: step 5 GTSAM}
\end{subfigure}
\newline
\begin{subfigure}{.45\linewidth}
  \centering
     \vspace*{8pt}
  \includegraphics[width=1.0\linewidth]{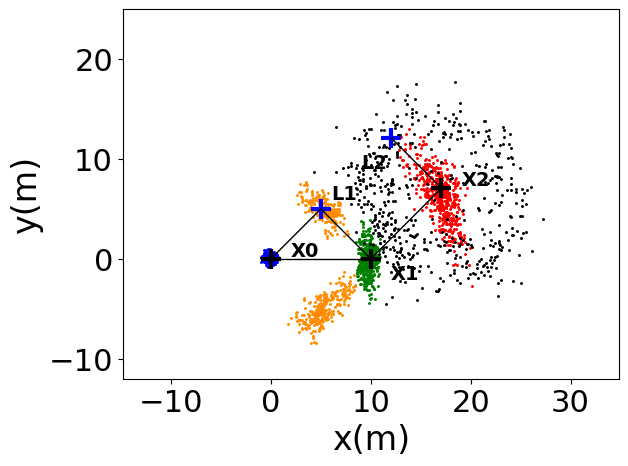}
  \vspace*{-12pt}
  \subcaption{Step 2 ({mm-iSAM})}
  \label{fig: step 2 Caesar}
\end{subfigure}
\begin{subfigure}{.45\linewidth}
  \centering
     \vspace*{8pt}
  \includegraphics[width=1.0\linewidth]{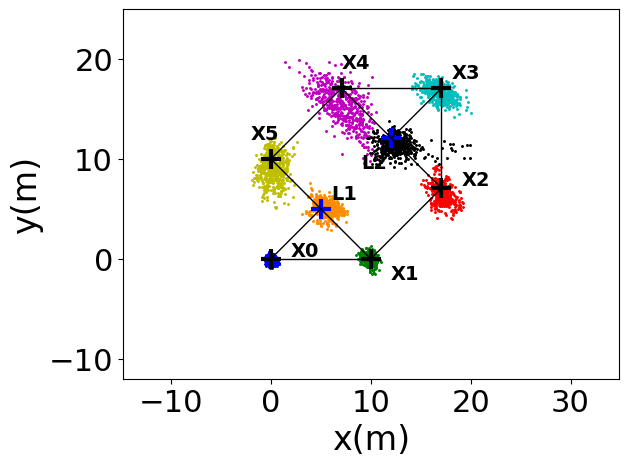}
  \vspace*{-12pt}
  \subcaption{Step 5 ({mm-iSAM})}
  \label{fig: step 5 Caesar}
\end{subfigure}
\newline
\begin{subfigure}{.45\linewidth}
  \centering
     \vspace*{8pt}
  \includegraphics[width=1.0\linewidth]{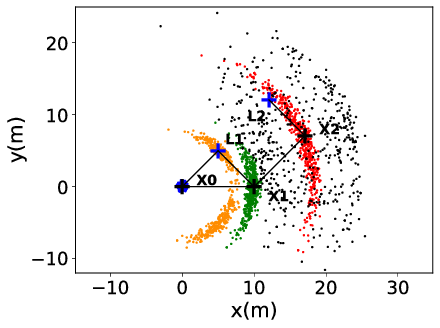}
  \vspace*{-12pt}
  \subcaption{Step 2 ({dynesty})}
  \label{fig: step 2 dynesty}
\end{subfigure}
\begin{subfigure}{.45\linewidth}
  \centering
     \vspace*{8pt}
  \includegraphics[width=1.0\linewidth]{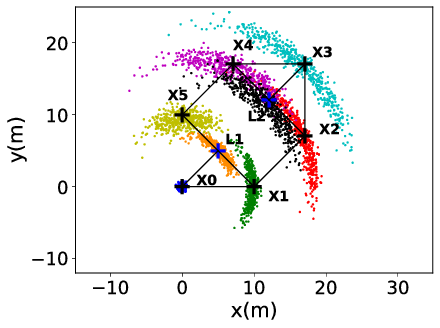}
  \vspace*{-12pt}
  \subcaption{Step 5 ({dynesty})}
  \label{fig: step 5 dynesty}
\end{subfigure}
\caption{Scatter plot of samples at time step 2 and time step 5, for four different methods: (a,b) NF-iSAM; (c,d) iSAM2; (e,f) mm-iSAM; (g,h) dynesty.  At time step 2, the robot ($X$) makes two movements and three distance measurements in which two landmarks ($L$) are spotted; At time step 5, the robot makes five movements and six distance measurements. Landmark and robot ground truth positions are marked by "+" while ground truth trajectory and range detection are plotted by solid lines.}
\label{fig: step 2 and 5}
\end{figure}

Fig. \ref{fig: step 2 and 5} shows samples of the posterior density at time step 2 and time step 5 on the $x$-$y$ plane (see Fig. \ref{fig: fig 1 illustration} for the Bayes tree at time step 2). 6 knots and 500 training samples were used in NF-iSAM for this experiment.
Since Gaussian noises of pose transformations are defined in the Lie algebra of $\mathrm{SE}(2)$, the resultant robot position samples on $x$-$y$ are distributed with a banana shape \cite{barfoot2014associating,long2013banana}. In Fig. \ref{fig: step 2 flow}, landmark $L_1$ has been spotted twice from different robot poses so it has two probable locations. The distance to landmark $L_2$ is only measured once so its density is supposed to be circular; however, as the robot pose ($X_2$) detecting it has a banana-shaped density, the $L_2$ posterior turns out to be a mix of a banana and a ring shape. Both NF-iSAM and {dynesty} are able to capture the correct shape. As the robot moves to $X_5$, where three distinct range measurements have been made to each landmark, the landmarks can be located by trilateration. As shown in Fig. \ref{fig: step 5 flow}, samples of both landmarks converge to be unimodal, which demonstrates the smoothing capacity of our algorithm. It should be mentioned that when running {iSAM2} on this problem for time step 0 and time step 2, the information matrix is singular due to landmarks not being sufficiently constrained by the measurements. In order to obtain a solution with {iSAM2}, we added Gaussian priors with a large variance on the landmarks when using {iSAM2}. The mean values for these artificial priors are set to the ground truth, which would not be known in practice; but even with the extra information, {iSAM2} still performs worse than the other methods (note that the linearly distributed samples in Fig.~\ref{fig: step 2 GTSAM} are with $L_2$).

\setlength\intextsep{0 pt}

\begin{figure}[t]\vspace*{3mm}
\centering
  \centering
  \includegraphics[width=0.95\linewidth]{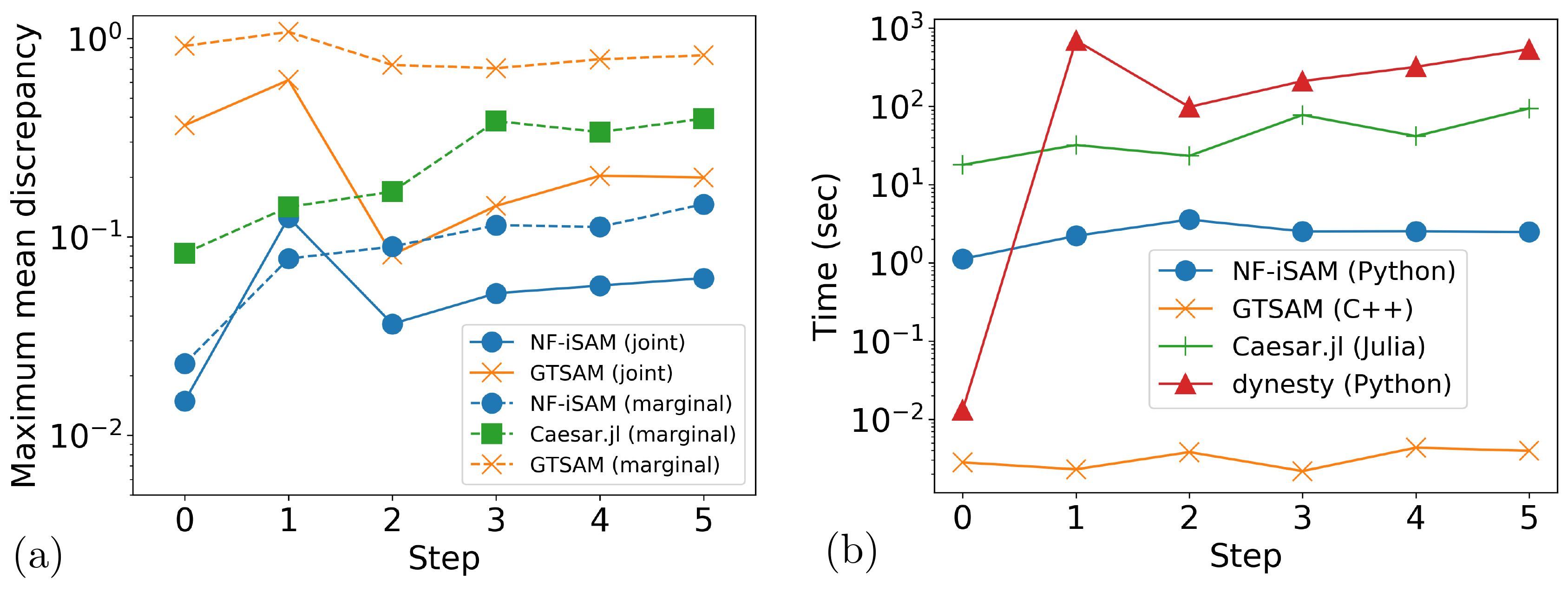}
 \vspace{-0.2cm}
\caption{Comparison against different approaches (iSAM2 provided by GTSAM and mm-iSAM provided by Caesar.jl): (a) MMD to {dynesty} joint distribution samples and averaged MMD over all variables to {dynesty} marginal distribution samples and (b) time to solve each step. Note that the vertical axis has logarithmic scale.}
\label{fig: accuracy and timing comparison}
\vspace{+0.2cm}
  \centering
  \includegraphics[width=.95\linewidth]{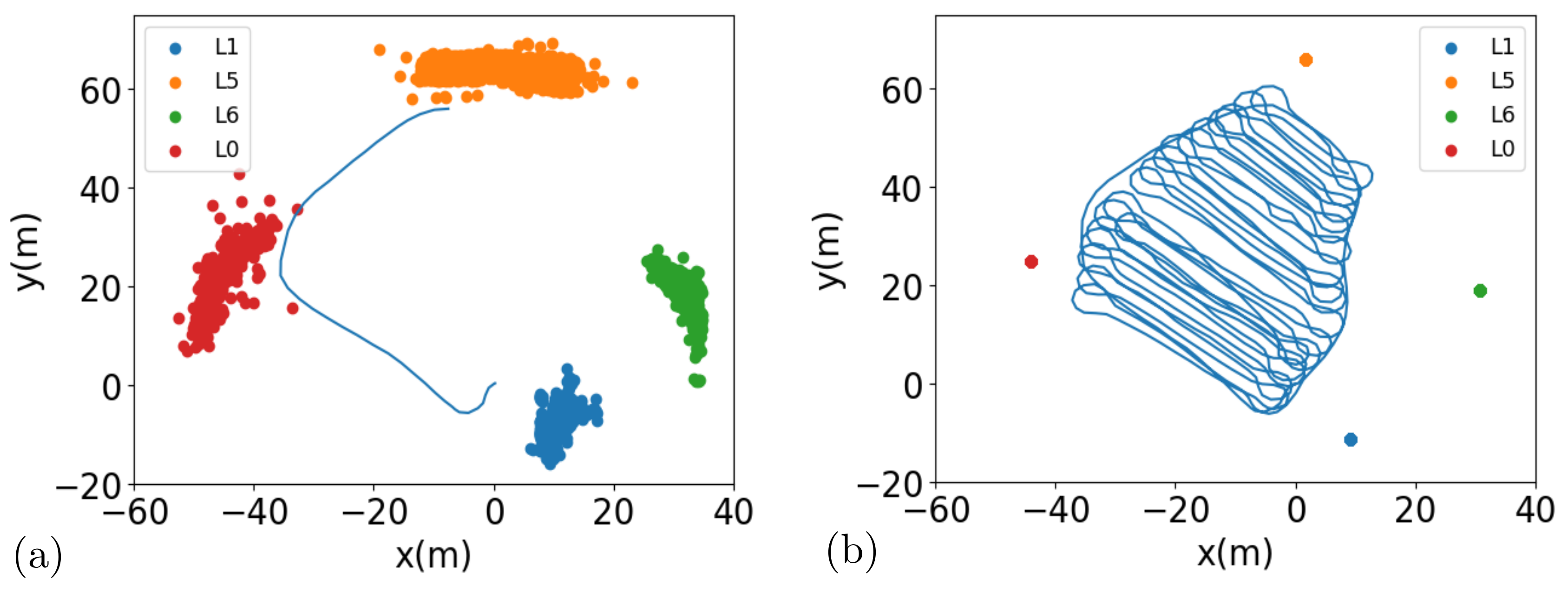}
\vspace{-0.2cm}
\caption{Posterior estimation of {Plaza1} dataset obtained by NF-iSAM: landmark samples and robot trajectory (average of robot samples). (a) Early stage (step 709 in  Fig. \ref{fig: plaza time and traj}) and (b) full trajectory (step 9368 in Fig. \ref{fig: plaza time and traj})}
\label{fig: plaza1 samples}
\vspace{+0.2cm}
\end{figure}

In the SLAM literature, root-mean-square error (RMSE) is usually employed to evaluate how far the MAP solution is from the ground truth. Here, since we aim to infer the full posterior distribution instead of a point estimate, the maximum mean discrepancy (MMD) \cite{gretton2012kernel} is a more reasonable choice. Given samples from two densities, MMD is a metric to evaluate how far the two distributions are apart. Therefore, a lower MMD from a solution to the {dynesty} solution indicates a more ``accurate" approximation of the posterior. As shown in Fig. \ref{fig: accuracy and timing comparison}(a), our solution is much closer to {dynesty} compared with others. Another important feature of NF-iSAM is that, similar to iSAM2, the output samples are from the joint posterior distribution. On the contrary, mm-iSAM can only compute marginal posterior distributions. As a result, in Fig. \ref{fig: accuracy and timing comparison}(a), it cannot be compared with the reference solution of the joint distribution. While having the best accuracy, as shown in Fig.\ref{fig: accuracy and timing comparison}(b), NF-iSAM (despite our Python implementation) is still comparable to mm-iSAM in terms of speed.

\setlength\intextsep{0 pt}

\subsection{The Plaza Dataset}
We also evaluate the scalability and computational cost of NF-iSAM on a larger real SLAM dataset.
The Plaza dataset provides time stamped range and odometry measurements ($\delta x, \delta \theta$) of a vehicle moving in a planar environment~\cite{djugash2009navigating}. One of its sequences, {Plaza1}, is available in the {GTSAM} software distribution. There are four unknown landmarks in the sequence. The range-only dataset is challenging for the state-of-the-art SLAM techniques (e.g., iSAM2) that rely on Gaussian approximation obtained by linearization around a posterior mode. At the early stage, landmarks are not completely constrained by the range measurements, which makes the information matrix to be singular or ill-conditioned, and the posterior to be multi-modal. {iSAM2} does not converge to a solution until the robot has moved hundreds of steps and good initial values are available. As shown in Fig.~\ref{fig: plaza1 samples}, our approach can solve for posterior distributions in the sequence from beginning to end. 10 knots and 1000 training samples were used in NF-iSAM for this experiment. The result shows that we can infer ``highly non-Gaussian'' distributions that arise during the early stage. 

\begin{figure}[t]\vspace*{3mm}
      \centering
      \includegraphics[width=.95\linewidth]{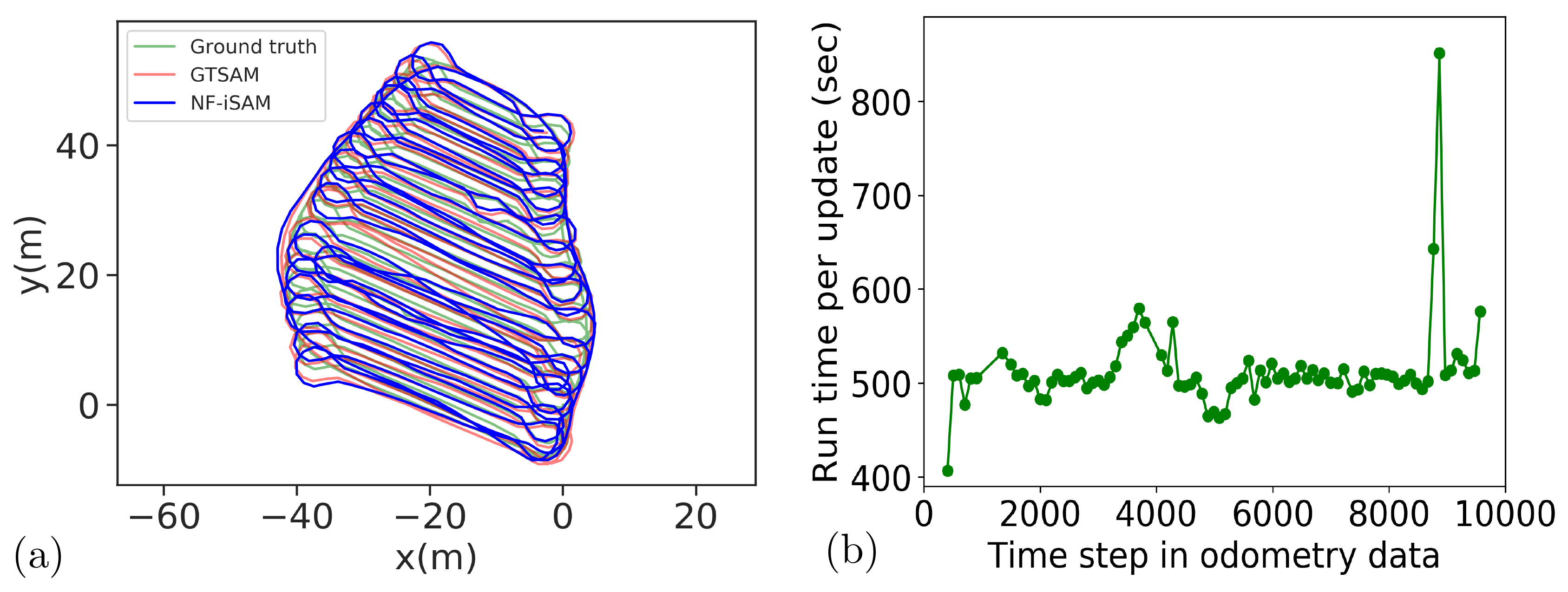}
    \vspace{-.2cm}
\caption{(a) Estimated full trajectories (iSAM2 provided by GTSAM), and (b) run time of NF-iSAM on Plaza1 dataset. Evo with Umeyama alignment was used to align the estimates with the ground truth\cite{grupp2017evo}.}
\label{fig: plaza time and traj}
\end{figure}

Fig.~\ref{fig: plaza time and traj}(a) presents estimated full trajectories by NF-iSAM and iSAM2 and the ground truth. The absolute trajectory error (ATE) averaged over the pose number in the NF-iSAM solution is 1.39 meters, and that in the iSAM2 solution is 0.99 meters. The error in part is due to the fact that not all the range and odometry measurements are added to the factor graph. To reduce the computational burden for this complex problem, odometry measurements with small displacement ($<10^{-2}$ meters) and rotation ($<10^{-3}$ radians) are skipped since we assume those measurements were taken when the robot stayed in-place. We designate 10 admitted odometry measurements as a batch. Only the range measurements taken at the end of a batch are added to our factor graph. Thus, the factor graph we solved has been simplified compared with that admitting all raw data. Note that the trajectory from NF-iSAM is obtained by computing the mean of robot pose samples (i.e., posterior mean vs.\ posterior mode sought by iSAM2), partially leading to the increased ATE. Although NF-iSAM aims at inferring the posterior distribution instead of point estimates, our trajectory estimate still resembles the ground truth well; furthermore, NF-iSAM can adequately represent the non-Gaussian joint posterior distribution (Fig.~\ref{fig: plaza1 samples}(a)).

 Fig.~\ref{fig: plaza time and traj}(b) shows the run time of our algorithm. We set our solver to perform incremental inference every 100 time steps (i.e., 10 batches). The run time per update is quite even as the robot proceeds, reflecting small and consistent sizes of cliques on the sub-tree for incremental inference (Section \ref{sec: incremental update}). The spike occurring around time step 9000 is caused by excessive range measurements at that point, which leads to larger clique sizes on the sub-tree of that incremental update. This mostly constant run time demonstrates that, under the non-Gaussian setting, we can still exploit sparsity of SLAM factor graphs via the Bayes tree which iSAM2 \cite{kaess2012isam2} exploited for Gaussian cases. 

\setlength\intextsep{0 pt}

\section{Conclusion and Future Work}
We presented a novel algorithm, NF-iSAM, that provides a promising foundation for addressing non-Gaussian inference problems encountered in SLAM due to non-linear measurement models and non-Gaussian (e.g., multi-modal) factors. NF-iSAM utilizes the Bayes tree coupled with neural networks to achieve efficient incremental updates to the joint posterior distribution for non-Gaussian factor graphs. We demonstrated the advantages of the approach over alternative state-of-the-art Gaussian and non-Gaussian SLAM techniques, with a synthetic dataset and a real dataset. 

We conclude the paper by noting that by restricting normalizing flow in NF-iSAM to affine transformations, our approach recovers iSAM2 as a special case. This shows that NF-iSAM is a ``canonical" extension of iSAM2 to the non-Gaussian setting, thus warranting further research as a promising algorithmic framework.

Future work will focus on additional experimental evaluation, on exploring the speed-accuracy trade-off, and on extending the range of applications, including:
(1) experimentally evaluate the algorithm with more sources of non-Gaussianity like outlier rejection;
(2) optimizing the normalizing flows, via neural network structures, knots of splines, and optimization method, to achieve better efficiency while maintaining accuracy;
(3)  utilizing faster incremental update strategies on the Bayes tree such as marginalization operations and variable elimination ordering with heuristics\cite{dellaert2012factor,Fourie2020wafr};
(4)   improving the treatment of orientation. Recently, more sophisticated treatments for orientation have been proposed for normalizing flows, which may enable customized parameterizations for robotics applications~\cite{rezende2020normalizing}; and
(5) generalizing the fixed spline-based parameterization to an adaptive parameterization that can switch between Gaussian and non-Gaussian SLAM problems.

\bibliographystyle{IEEEtran}
\bibliography{ref.bib}
\end{document}